\documentclass[conference]{IEEEtran}

\usepackage[T1]{fontenc}
\usepackage{amsmath,amssymb}
\usepackage{cite}
\usepackage{graphicx}
\usepackage{tikz-cd}
\usepackage{url}
\usepackage{array}
\usepackage{listings}
\usepackage[hidelinks]{hyperref}

\lstdefinestyle{promptstyle}{%
basicstyle=\ttfamily\scriptsize,
breaklines=true,
breakatwhitespace=true,
columns=fullflexible,
frame=single,
captionpos=b
}

\title{From Entity Mentions to Tone: An LLM-Based Pipeline for Media Bias Analysis}

\author{%
\IEEEauthorblockN{Klesti Hoxha and Olti Qirici}
\IEEEauthorblockA{%
Department of Informatics\\
Faculty of Natural Sciences\\
University of Tirana\\
Tirana, Albania\\
\{klesti.hoxha, olti.qirici\}@unitir.edu.al}
\thanks{Author's accepted manuscript of K. Hoxha and O. Qirici, ``From Entity Mentions to Tone: An LLM-Based Pipeline for Media Bias Analysis,'' in \emph{2026 18th International Conference on Electronics, Computers and Artificial Intelligence (ECAI)}, 2026, pp. 1--8, doi: \href{https://doi.org/10.1109/ECAI69016.2026.11613638}{10.1109/ECAI69016.2026.11613638}. \copyright~2026 IEEE. Personal use of this material is permitted. Permission from IEEE must be obtained for all other uses, in any current or future media, including reprinting/republishing this material for advertising or promotional purposes, creating new collective works, for resale or redistribution to servers or lists, or reuse of any copyrighted component of this work in other works.}
}

\begin{document}
\renewcommand\IEEEkeywordsname{Keywords}
\IEEEoverridecommandlockouts
\IEEEpubid{\makebox[\columnwidth]{\copyright2026 IEEE\hfill} \hspace{\columnsep}\makebox[\columnwidth]{}}

\maketitle

\begin{abstract}
This paper presents a pipeline for analyzing media bias and framing in online news. The pipeline groups articles into topics and events, adds named-entity and sentiment annotations, and compares news sources through people mentions, source-level tone, and event-level coverage patterns. We apply it to 8,358 Albanian news articles collected from GDELT and compare the resulting annotations with GDELT's automated annotations. The results show moderate agreement for sentiment and entity extraction, as well as additional person-entity pairs that can potentially support the bias analysis. We compare two annotation prompts and find that stricter sentiment-validation rules remove label-score inconsistencies but increase execution time and reduce annotation coverage. Based on these results, the simpler prompt is used for the rest of the analysis. We have provided sample analysis on source-level framing profiles, person-level tone differences across sources, and event-level gatekeeping and coverage indicators. These outputs show how the same news collection can be used to examine what sources cover, how they describe public figures, and where coverage is concentrated. The approach is particularly useful in settings where manually verified datasets or specialized language tools are limited.
\end{abstract}
\vspace{0.6em}


\section{Introduction}

Online news providers have become the default way for many people to be informed. They can be accessed directly or, quite often, through social media platforms. In both cases, recommender-system algorithms \cite{raza_news_2022} boost specific news items or topics \cite{hoxha_real_2026}, making readers vulnerable to disinformation and selective representations of events or facts \cite{wang_media_2025}. Selection may focus on individuals, organizations, or specific topics covered in the news \cite{wang_media_2025}, while framing can be related to the tone used when covering people, organizations, or topics in specific news items \cite{r_sentinel_2025}.

Traditionally, this problem has been addressed through manual investigations of news providers by communication experts \cite{castillo-campos_automated_2025}. However, this process is heavyweight, costly, and itself prone to bias. Furthermore, it is difficult to apply continuously at the level of individual media providers or news articles. 

As an alternative, machine-learning algorithms, such as classifiers, have been used to identify bias or framing in media content \cite{wessel_llm-based_2025}. However, these methods also rely on language-specific training data, which are usually created by human annotators.

To address gatekeeping and reduce disinformation \cite{lazer_science_2018} in light of recent advances in news dissemination technologies, reliable and fact-centered tools are needed to continuously assess bias and framing at the level of individual news providers and articles. Such tools are able to enhance news aggregator applications by enabling the visualization of these assessments.

This need is especially visible in smaller language settings such as Albanian. News is published by many outlets with different editorial positions, but there are fewer ready-made resources for large-scale entity recognition, sentiment analysis, and source comparison. As a result, even basic questions can be difficult to answer consistently: which sources covered an event, which public figures were mentioned, and whether the tone differed across outlets.

For this reason, media-bias analysis should not rely on a single score. It should separate several related signals: source-level tendencies, person-level tone, and event-level coverage. Source-level summaries help show whether a provider is generally more favorable or critical. Person-level summaries show how public figures are described by different outlets. Event-level indicators show how widely an event is covered and where coverage is concentrated.

Recent advances in multilingual LLMs have enabled accurate and reliable approaches to NLP tasks such as named-entity recognition (NER) and sentiment analysis \cite{qin_large_2026}. Both tasks are strongly related to the automated monitoring of bias and framing in news articles. In this work, we introduce a language-independent pipeline that continuously monitors media bias and framing by using LLMs to track mentions and tone. Beyond a static evaluation of a news provider's bias toward individuals or organizations, the pipeline compares coverage of the same event across different news outlets to identify shifts in tone and visibility. We test this system in a low-resource media environment, showing how local open-source language models can support structured sentiment and entity analysis without requiring prior language-specific adaptation.

This work makes four main contributions. First, it implements a pipeline, built with Kedro and a local Gemma model, for studying bias, framing, and gatekeeping in Albanian news. Second, it compares two annotation prompts and shows that stricter validation removes label-score inconsistencies but also slows processing and reduces annotation coverage. Third, it compares the resulting annotations with GDELT outputs on 8,358 Albanian news articles. Fourth, it uses the annotated dataset to provide a general framework for evaluating source-level framing profiles, person-level tone differences, and event-level coverage indicators across 124 news sources, establishing a reproducible blueprint applicable to other low-resource language news environments.

The rest of the paper follows this structure. We first define the bias and framing categories used in the study and then review related work. Next, we describe the dataset, prompts, and processing pipeline. We then evaluate the annotations and present the source-, person-, and event-level analyses. The paper ends with a discussion of the main limitations of the approach and potential directions for future work.

\section{Bias and Framing Categories}

From the reviewed literature, we identified several recurring categories of media bias and framing. These categories distinguish between what news outlets choose to cover, how they describe it, and how much attention they allocate to specific actors or events.

Media bias is commonly treated as a deviation from neutrality \cite{wang_media_2025}. Table \ref{table:bias_categories} summarizes the bias categories considered in this work.

\begin{table}[hb]
\renewcommand{\arraystretch}{1.4} % Adds padding for academic readability
\caption{Media Bias Categories}
\label{table:bias_categories}
\centering
\scriptsize
\begin{tabular}{>{\raggedright\arraybackslash}m{1.8cm}p{4.0cm}p{1.4cm}}
    \hline
    \textbf{Category} & \textbf{Description} & \textbf{Reference} \\
    \hline
    Selection & Editorial ``gatekeeping''; choosing what is deemed newsworthy or omitting specific details. & \cite{wang_media_2025} \\
    Tone & The emotional ``lean'' or sentiment valence applied to specific actors or events. & \cite{tohidi_rethinking_2025} \\
    Reporting/Source & The long-term reputation, systemic lean, and ideological history of a news domain. & \cite{ronnback_automatic_2025} \\
    Depth & Uneven coverage space or word count dedicated to one side of a story. & \cite{r_sentinel_2025} \\
    \hline
\end{tabular}
\end{table}

Media framing captures how a narrative is structured once an event is selected for coverage. Table \ref{table:framing_categories} lists the framing categories used to characterize these narrative choices.

\begin{table}[htb]
    \renewcommand{\arraystretch}{1.4} % Adds padding for academic readability
    \caption{Media Framing Categories}
    \label{table:framing_categories}
    \centering
    \scriptsize
    \begin{tabular}{>{\raggedright\arraybackslash}m{1.8cm}>{\raggedright\arraybackslash}p{4.0cm}p{1.4cm}}
    \hline
    \textbf{Category} & \textbf{Description} & \textbf{Reference} \\
    \hline
    Selection & Structural visibility of narrative elements; what information is presented or omitted. & \cite{wang_media_2025,tohidi_rethinking_2025} \\
    Tone & The emotional undertone, spin, or evaluative posture applied to the news narrative. & \cite{r_sentinel_2025,tohidi_rethinking_2025} \\
    Emphasis & The relative importance and priority given to specific themes or viewpoints within a text. & \cite{wang_media_2025} \\
    Targeted & Entity-specific framing shifts depending on the politician or actor involved. & \cite{elbouanani_analyzing_2025} \\

    \hline
    \end{tabular}
\end{table}

\section{Related Work}

Several recent studies have used LLMs to detect bias and framing in news outlets. 

Wang et al. \cite{wang_media_2025} created a real-time dashboard that visualizes bias metrics such as tone, topic, and political lean. LLMs are used to quantify and categorize these metrics. The user interface allows users to browse individual events and coverage details, while also providing aggregated bias metrics at the publisher level. The developed system was evaluated with communication experts and crowdsourced participants. The results were promising, showing a high statistical correlation with human-based annotations.

In another study, Kumar et al. \cite{r_sentinel_2025} created a framework and application that measure possible bias at the level of individual stories or events. It allows users to compare news stories side by side, highlighting possible polarization in their framing. The authors use LLMs to perform sentiment analysis and emphasize their benefits for deeper, context-based sentiment detection. LLMs are also used to compare coverage across news providers reporting on the same story.

In earlier pre-LLM work, Ye and Skiena \cite{ye_mediarank_2019} computed source-related rankings for 50,000 news sources globally. Their ranking framework included reporting bias, peer reputation, and social-media-based popularity. The results were evaluated against independent expert scorecards and showed good agreement.

Similarly, R\"onnback et al. \cite{ronnback_automatic_2025} used machine learning to perform large-scale labeling and classification of large datasets of news providers (domains). The outputs of their system were human-interpretable. Training was based on the GDELT dataset \cite{leetaru2013gdelt}, while evaluation was conducted using a human-labeled dataset. They also included an LLM baseline, but it performed worse than a non-LLM neural network.

LLMs have also been used for dataset augmentation and synthetic news generation. Wessel \cite{wessel_llm-based_2025} used LLMs to augment a dataset of bias-indicating keywords by replacing keywords with generated alternatives. This approach aims to reduce the effect of ``spurious cues'' in machine-learning models. The author argues that the resulting model is less dependent on the context of the observed keywords. Tohidi et al. \cite{tohidi_rethinking_2025}, on the other hand, used LLMs to demonstrate that generated biased news can affect public opinion. Their results showed that the way news is presented to readers can influence their perceptions, even when the underlying facts remain unchanged. This further indicates the need for better ways to address bias and framing when combating disinformation.

Despite the successful use of LLMs for detecting bias and framing, LLMs have also been shown to carry biases. Elbouanani et al. \cite{elbouanani_analyzing_2025} found experimentally that tone (sentiment) classification differs for identical phrases when the political targets in those phrases are substituted. As a mitigation strategy, they show that replacing politician names with fictitious ones reduces bias during sentiment classification. In this way, the context in which those politicians are mentioned carries more weight in the sentiment classification.

\section{Method}

\begin{figure}[htbp]
    \centering
    \includegraphics[height=1.2\columnwidth,keepaspectratio]{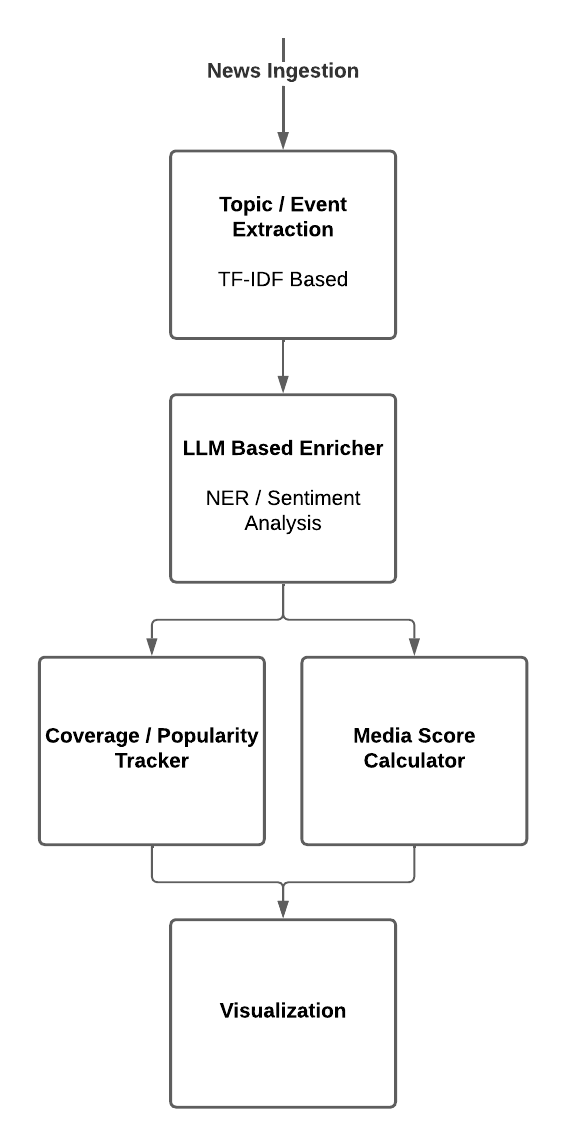}
    \caption{Overview of the proposed media-bias analysis pipeline.}
    \label{fig:media_bias_pipeline}
\end{figure}

In this work, we develop a media-bias analysis pipeline that tracks mentions of people in news articles and measures their sentiment polarization, or tone (Fig. \ref{fig:media_bias_pipeline}). The tracked news articles are written in Albanian, a low-resource language. 

To the best of our knowledge, there is no widely used state-of-the-art NER toolkit for Albanian, and the same holds for sentiment analysis. Most previous work consists of scholarly experiments involving the creation of small-scale, human-annotated datasets \cite{kadriu_human-annotated_2022,barolli_ner_2025} or fine-tuned language models such as XLM-RoBERTa \cite{nuci_roberta_2024}. Therefore, using LLMs to tackle these tasks may offer practical benefits.

For our experiments, we used a dataset of 8,358 Albanian news articles from GDELT \cite{leetaru2013gdelt}, published in April 2026 (Table \ref{table:dataset_descriptive_stats}). To detect potential bias toward specific topics or events, the incoming news items are grouped into topics using a simple TF-IDF and k-means approach. We then identify events within these topics using TF-IDF in combination with cosine similarity and time windows. Each news item, identified by a unique URL, is enriched with NER and sentiment-analysis outputs using Gemma 4\footnote{\url{https://ai.google.dev/gemma}}, a local LLM.

\begin{table}[htbp]
\renewcommand{\arraystretch}{1.4} % Adds padding for academic readability
\caption{News Dataset Details}
\label{table:dataset_descriptive_stats}
\centering
\scriptsize
    \begin{tabular}{>{\raggedright\arraybackslash}m{3.8cm}p{3.2cm}}
    \hline
    \textbf{Metric} & \textbf{Value} \\
    \hline
    Articles & 8,358 \\
    Unique news sources & 124 \\
    Topic clusters & 50 \\
    Event clusters & 3,551 \\
    Multi-source events & 1,340 \\
    Articles per source (min/max/mean/median) & 1 / 667 / 67.4 / 23 \\
    Articles per topic (min/max/mean/median) & 14 / 699 / 167.16 / 127.50 \\
    Articles per event (min/max/mean/median) & 1 / 48 / 2.35 / 1 \\
    Article text length chars (mean/median) & 1622 / 1379 \\
    \hline
    \end{tabular}
\end{table}

The pipeline was developed using the Kedro\footnote{\url{https://kedro.org/}} Python framework. It measures bias and framing by tracking tone variation in topic and event coverage, as well as tonal bias in mentions of people. It also computes possible gatekeeping indicators by analyzing which people, topics, or events are repeatedly covered or omitted across news sources.

\section{Experiments and Results}

In this section, we describe the experiments and their results. Our aim is to demonstrate the feasibility of implementing such a pipeline by presenting potential use cases.

\subsection{NER and Sentiment Enrichment}

Our experiments were run on a cloud instance with an NVIDIA A10 GPU (24 GB), 30 vCPUs, and 200 GiB of RAM. We relied on the local LLM \textit{Gemma 4} for NER and sentiment-analysis enrichment. The model temperature was set to 0.0 to obtain deterministic outputs, and generation was limited to 250 tokens. Requests used a 90-second timeout with one retry in case of failure. The prompts instructed the model to extract sentiment and NER data from the full text of each news article and return the output in JSON format.

Listings \ref{lst:prompt_v1} and \ref{lst:prompt_v2} show the two prompts used for NER and sentiment enrichment. The output of the first version showed inconsistencies between the sentiment labels (positive, negative, neutral) and their corresponding scores. The second version addresses this issue by adding explicit validation rules.

\begin{lstlisting}[style=promptstyle,caption={Prompt v1 for NER and sentiment enrichment},label={lst:prompt_v1}]
You are an information extraction assistant.
Return JSON only with this exact schema:
{
  "sentiment": {"label": "positive|neutral|negative", "score": number between -1 and 1},
  "entities": {
    "PER": [{"original": string, "canonical_en": string}],
    "ORG": [{"original": string, "canonical_en": string}],
    "LOC": [{"original": string, "canonical_en": string}]
  }
}.
Rules: canonical_en must be an English canonical form, keep original as text surface form from article, omit uncertain entities, no extra keys, no markdown.

ARTICLE:
{article_text}
\end{lstlisting}

\begin{lstlisting}[style=promptstyle,caption={Prompt v2 for NER and sentiment enrichment},label={lst:prompt_v2}]
You are an information extraction assistant.
Return JSON only with this exact schema:
{
  "sentiment": {"label": "positive|neutral|negative", "score": number between -1 and 1},
  "entities": {
    "PER": [{"original": string, "canonical_en": string}],
    "ORG": [{"original": string, "canonical_en": string}],
    "LOC": [{"original": string, "canonical_en": string}]
  }
}.
Rules:
1. Sentiment score must range from -1.0 (very negative) to 1.0 (very positive).
2. Label MUST match score: use 'negative' if score < -0.3, 'positive' if score > 0.3, 'neutral' otherwise.
3. canonical_en must be an English canonical form, keep original as text surface form from article.
4. Omit uncertain entities, no extra keys, no markdown, and no reasoning text.

ARTICLE:
{article_text}
\end{lstlisting}

Prompt v2 was slower and produced fewer labeled articles for both sentiment and NER (Table \ref{table:llm_versions}). The inconsistency rate between sentiment labels and scores in the first version was 8.53\%. The gain in consistency did not justify the doubled execution time, and the remaining inconsistencies can be addressed in a cleaning step.

\begin{table}[htbp]
\renewcommand{\arraystretch}{1.4} % Adds padding for academic readability
\caption{LLM Enrichment Versions Comparison}
\label{table:llm_versions}
\centering
\scriptsize
    \begin{tabular}{>{\raggedright\arraybackslash}m{3.6cm}p{1.5cm}p{1.5cm}}
    \hline
    \textbf{Metric} & \textbf{Prompt v1} & \textbf{Prompt v2} \\
    \hline
    Sentiment Labeled Articles & 51.75\% & 37.20\% \\
    Inconsistent Sentiment Label / Score & 8.53\% & 0.0\% \\
    NER-Labeled Articles (any entity) & 39.97\% & 27.20\% \\
    Articles with Person Entities (PER) & 8.25\% & 5.77\% \\
    \hline
    \end{tabular}
\end{table}

In the absence of a gold-standard test dataset, we evaluated the quality of the LLM-based annotations by measuring their agreement with the annotations provided by GDELT. These annotations are also generated by an automated NLP pipeline. The NER evaluation was based only on the PERSON category because our pipeline focuses on mentions of people. For this evaluation, we considered only articles labeled by both GDELT and the LLM-based annotator. We used the following evaluation metrics.

For sentiment, we report the number of comparable articles and the label agreement rate, defined as the percentage of articles where the LLM and GDELT sentiment labels match. For NER, we compare URL--entity-type pairs and report three macro-averaged agreement rates: one from the LLM side, one from the GDELT side, and a balanced score that averages both perspectives. 

Overall, the agreement rates in Table \ref{table:sentiment_eval} and Table \ref{table:ner_eval} show moderate alignment between the LLM-based annotations and the GDELT annotations. For sentiment, Prompt v2 achieves a slightly higher label agreement rate than Prompt v1. For NER, the LLM identifies more person-entity pairs than GDELT. At the same time, the high GDELT-side agreement shows that entities identified by GDELT are also found by the LLM to a large degree. 

Rather than contradicting the existing data, the LLM actually validates it. It recovers the majority of GDELT's person mentions while successfully identifying additional ones the source overlooked. These additional pairs should not be treated as automatically correct without manual validation, but when lacking a better alternative, they can still be used for media-bias analysis because missed person mentions can reduce the quality of source-level and person-level tone comparisons. Because the analysis focuses on people mentioned in the news, less frequent mentions can still be manually verified in the later stages of the pipeline.

\begin{table}[htbp]
\renewcommand{\arraystretch}{1.4} % Adds padding for academic readability
\caption{Sentiment Evaluation Against GDELT}
\label{table:sentiment_eval}
\centering
\scriptsize
    \begin{tabular}{>{\raggedright\arraybackslash}m{3.6cm}p{1.5cm}p{1.5cm}}
    \hline
    \textbf{Metric} & \textbf{Prompt v1} & \textbf{Prompt v2} \\
    \hline
    Comparable articles & 3,612 & 3,109 \\
    Label agreement rate & 58.72\% & 62.34\% \\
    \hline
    \end{tabular}
\end{table}

\begin{table}[htbp]
\renewcommand{\arraystretch}{1.4} % Adds padding for academic readability
\caption{People-Only NER Agreement Rates Against GDELT}
\label{table:ner_eval}
\centering
\scriptsize
    \begin{tabular}{>{\raggedright\arraybackslash}m{3.6cm}p{1.5cm}p{1.5cm}}
    \hline
    \textbf{Metric} & \textbf{Prompt v1} & \textbf{Prompt v2} \\
    \hline
    Evaluable pairs & 690 & 483 \\
    Macro LLM agreement rate & 36.83\% & 36.12\% \\
    Macro GDELT agreement rate & 71.46\% & 70.41\% \\
    Macro balanced agreement rate & 45.30\% & 44.44\% \\
    \hline
    \end{tabular}
\end{table}

While the agreement rates in Table~\ref{table:ner_eval} measure overlap from both the LLM and GDELT perspectives for person-name identification, we also report precision, recall, and F1 scores to summarize the same comparison in a more standard information extraction format. The results are very similar across prompt versions. Recall is higher than precision because the LLM-based person recognizer identifies most names also found in the GDELT dataset while extracting additional names.

\begin{table}[htbp]
\renewcommand{\arraystretch}{1.4} % Adds padding for academic readability
\caption{People-Only NER Precision and Recall Against GDELT}
\label{table:ner_performance_by_version}
\centering
\scriptsize
\resizebox{\columnwidth}{!}{%
    \begin{tabular}{lrrrrr}
    \hline
    \textbf{Version} & \textbf{LLM people} & \textbf{GDELT people} & \textbf{Precision} & \textbf{Recall} & \textbf{F1} \\
    \hline
    v1 & 1860 & 799 & 30.2\% & 70.3\% & 42.3\% \\
    v2 & 1309 & 560 & 29.8\% & 69.6\% & 41.7\% \\
    \hline
    \end{tabular}%
}
\end{table}

The evaluation results suggest that, for low-resource languages such as Albanian, LLM-based annotation can provide a useful starting point when no gold-standard dataset or well-trained alternative tools are available for NER and sentiment extraction. GDELT is also based on automated annotation, so these results should be read as a comparison with an existing large-scale system rather than as a definitive evaluation. The agreement with GDELT suggests that the LLM outputs are generally consistent with an established reference, while also identifying additional entities that may be useful for bias analysis.

Furthermore, although Prompt v2 reduces inconsistencies between sentiment labels and scores, it does so at the cost of slower execution and does not substantially improve the evaluation results. Therefore, we decided to continue with the dataset labeled using Prompt v1.

\subsection{Media Bias and Framing Analysis}
\label{subsec:media_bias_framing_analysis}

Table \ref{table:aggregate_source_framing_excerpt} presents an excerpt from the source-level framing and bias profiles. The table summarizes how frequently each source uses positive, neutral, or negative framing, together with aggregate sentiment and bias measures. The reported measures are defined as follows.

\begin{itemize}

    \item \textbf{Pos. / Neu. / Neg.} --- share of the source's labeled articles classified as positive, neutral, and negative sentiment, respectively.

    \item \textbf{Mean score} --- average raw sentiment score across all labeled articles of the source; positive values indicate favorable coverage, while negative values indicate critical coverage.

    \item \textbf{Avg.\ bias} --- source-level framing bias, computed as the average of per-topic framing bias scores (positive rate minus negative rate for each topic), then averaged across all topics covered by the source.

    \item \textbf{Bias vs.\ corpus} --- deviation of the source's average bias from the corpus-wide baseline (baseline $= -0.192$ in this study). A positive value indicates the source frames topics more favorably than the corpus average; a negative value indicates systematically more critical coverage.
\end{itemize}

The results are broadly consistent with the generally perceived editorial positions of the these news sources in the Albanian news landscape. This analysis can help highlight differences in how news providers frame the same topics.

\begin{table*}[htbp]
\renewcommand{\arraystretch}{1.4} % Adds padding for academic readability
\caption{Aggregate Source Framing and Bias Profiles}
\label{table:aggregate_source_framing_excerpt}
\centering
\scriptsize
    \begin{tabular}{>{\raggedright\arraybackslash}p{2.6cm}rrrrrrrr}
    \hline
    \textbf{Source} & \textbf{Topics} & \textbf{Articles} & \textbf{Pos.} & \textbf{Neu.} & \textbf{Neg.} & \textbf{Mean score} & \textbf{Avg. bias} & \textbf{Bias vs. corpus} \\
  \hline
  tvkoha.tv & 2 & 10 & 22.7\% & 27.3\% & 50.0\% & +0.100 & -0.857 & -0.665 \\
  3shi.net & 4 & 16 & 40.7\% & 29.6\% & 29.6\% & +0.170 & +0.200 & +0.392 \\
  tv21.tv & 5 & 27 & 11.4\% & 50.0\% & 38.6\% & +0.177 & -0.554 & -0.362 \\
  kosovapress.com & 4 & 17 & 9.7\% & 25.8\% & 64.5\% & -0.203 & -0.542 & -0.350 \\
  rtsh.al & 1 & 33 & 38.5\% & 33.3\% & 28.2\% & +0.108 & +0.151 & +0.344 \\
  epokaere.com & 3 & 10 & 27.6\% & 34.5\% & 37.9\% & +0.324 & +0.111 & +0.303 \\
  sot.com.al & 18 & 122 & 9.6\% & 35.3\% & 55.1\% & +0.128 & -0.493 & -0.301 \\
  noa.al & 2 & 29 & 10.0\% & 36.7\% & 53.3\% & +0.123 & -0.475 & -0.283 \\
  vizionplus.tv & 10 & 59 & 12.0\% & 29.3\% & 58.7\% & -0.088 & -0.459 & -0.267 \\
  ina-online.net & 7 & 25 & 27.5\% & 42.5\% & 30.0\% & +0.170 & +0.071 & +0.264 \\
  \hline
    \end{tabular}
\par\vspace{0.6em}
\makebox[\textwidth][c]{%
\begin{minipage}{0.72\textwidth}
\raggedright\footnotesize Corpus baseline = -0.192; only sources with at least 10 sentiment labeled articles are included.
\end{minipage}%
}
\end{table*}

Table \ref{table:entity_tone_extremes_excerpt} shows an excerpt of a person-tone bias analysis. The table reveals how different news sources frame specific people in their coverage, showing whether sources systematically portray particular individuals more favorably or critically. For each person, only the lowest and highest tone-balance rows across different sources are displayed, illustrating the range of framing. The reported measures are defined as follows.

\begin{itemize}

    \item \textbf{Source} --- the news source that mentioned the person.

    \item \textbf{Mentioned articles} --- number of labeled articles from that source in which the person was mentioned.

    \item \textbf{Mean score} --- average raw sentiment score across all mentions of the person in that source's articles; positive values indicate favorable coverage, negative values indicate critical coverage.

    \item \textbf{Tone balance} --- framing balance for the person at that source, computed as positive rate minus negative rate across all mentions; ranges from $-1$ (uniformly negative) to $+1$ (uniformly positive).

    \item \textbf{Pos. / Neg.} --- share of mentions of the person in that source's articles classified as positive or negative sentiment, respectively.

\end{itemize}

This analysis can be used to compare how the same public figure is framed across news sources. For example, a news aggregator could surface cases where a person receives strongly positive coverage in one source and strongly negative coverage in another, helping readers identify possible framing differences around the same actor. While our practical testing focuses on well-known regional and international players to show effective local tracking, the basic extraction method works for any target and can easily handle any group of entities.

As noted in \cite{elbouanani_analyzing_2025}, pre-trained models may carry biases when processing mentions of politicians in news articles. However, the results can still serve as a useful baseline, and such biases can be mitigated through name anonymization or user-provided feedback.

\begin{table*}[htbp]
\renewcommand{\arraystretch}{1.4} % Adds padding for academic readability
\caption{Person-Tone Extremes by Person Across News Sources}
\label{table:entity_tone_extremes_excerpt}
\centering
\scriptsize
    \begin{tabular}{>{\raggedright\arraybackslash}p{3.0cm}>{\raggedright\arraybackslash}p{3.0cm}rrrrr}
    \hline
    \textbf{Person} & \textbf{Source} & \textbf{Mentioned articles} & \textbf{Mean score} & \textbf{Tone balance} & \textbf{Pos.} & \textbf{Neg.} \\
    \hline
    Benjamin Netanyahu & balkanweb.com & 5 & +0.000 & -0.800 & 0.0\% & 80.0\% \\
    Benjamin Netanyahu & panorama.com.al & 6 & +0.300 & -0.333 & 16.7\% & 50.0\% \\
    Donald Trump & opinion.al & 8 & -0.300 & -0.750 & 12.5\% & 87.5\% \\
    Donald Trump & lajmi.net & 6 & +0.400 & +0.500 & 50.0\% & 0.0\% \\
    Edi Rama & syri.net & 10 & +0.190 & -1.000 & 0.0\% & 100.0\% \\
    Edi Rama & kohajone.com & 8 & +0.600 & +0.250 & 62.5\% & 37.5\% \\
    Kreshnik Mujeci & top-channel.tv & 5 & -0.140 & -1.000 & 0.0\% & 100.0\% \\
    Kreshnik Mujeci & sot.com.al & 5 & -0.660 & -0.800 & 0.0\% & 80.0\% \\
    Sali Berisha & balkanweb.com & 11 & +0.455 & -0.818 & 9.1\% & 90.9\% \\
    Sali Berisha & syri.net & 6 & +0.667 & -0.500 & 16.7\% & 66.7\% \\
    \hline
    \end{tabular}
\par\vspace{0.6em}
\makebox[\textwidth][c]{%
\begin{minipage}{0.80\textwidth}
\raggedright\footnotesize Excerpt showing minimum and maximum tone balance per person; person-source pairs with $\geq 5$ mentions are included.
\end{minipage}%
}
\end{table*}

Table \ref{table:gatekeeping_coverage_indicators} presents event-level gatekeeping and coverage indicators. It shows how widely events are covered across news sources and how concentrated that coverage is. This approach is similar to those applied by Wang et al. \cite{wang_media_2025} and Kumar et al. \cite{r_sentinel_2025}. The reported measures are defined as follows.

\begin{itemize}

    \item \textbf{Event} --- event identifier, sorted by article count in descending order.

    \item \textbf{Articles} --- total number of articles assigned to that event.

    \item \textbf{Covering Sources} --- number of distinct news sources that covered the event.

    \item \textbf{Omitting Sources} --- number of active sources in the corpus that did not cover the event, computed as total active sources minus covering sources.

    \item \textbf{Gatekeeping Score} --- structural selectivity index defined as $1 - \frac{\text{covering sources}}{\text{total active sources}}$; higher values indicate narrower dissemination and stronger gatekeeping.

    \item \textbf{Framing Polarity} --- event-level sentiment balance computed as positive rate minus negative rate across labeled articles for that event; positive values indicate more favorable framing, negative values indicate more critical framing.

\end{itemize}

This analysis can be used to identify events that receive broad attention and events that are covered by only a small set of sources. For example, a news aggregator could flag events with high gatekeeping scores and compare their framing polarity across sources, helping readers notice possible gaps in coverage or differences in tone.

\begin{table*}[htbp]
\renewcommand{\arraystretch}{1.4} % Adds padding for academic readability
\caption{Gatekeeping and Coverage Indicators for Major Events}
\label{table:gatekeeping_coverage_indicators}
\centering
\scriptsize
    \begin{tabular}{>{\raggedright\arraybackslash}p{2.4cm}rrrrr}
    \hline
    \textbf{Event} & \textbf{Articles} & \textbf{Covering sources} & \textbf{Omitting sources} & \textbf{Gatekeeping score} & \textbf{Framing polarity} \\
    \hline
    Event 1 & 48 & 21 & 103 & 0.8306 & +0.0227 \\
    Event 2 & 46 & 35 & 89 & 0.7177 & +0.1818 \\
    Event 3 & 39 & 19 & 105 & 0.8468 & +0.0625 \\
    Event 4 & 37 & 22 & 102 & 0.8226 & --- \\
    Event 5 & 33 & 19 & 105 & 0.8468 & -1.0000 \\
    Event 6 & 33 & 10 & 114 & 0.9194 & +0.0588 \\
    Event 7 & 30 & 18 & 106 & 0.8548 & +1.0000 \\
    Event 8 & 29 & 23 & 101 & 0.8145 & -0.0625 \\
    Event 9 & 29 & 22 & 102 & 0.8226 & +0.7142 \\
    Event 10 & 29 & 14 & 110 & 0.8871 & --- \\
    \hline
    \end{tabular}
\par\vspace{0.6em}
\makebox[\textwidth][c]{%
\begin{minipage}{0.80\textwidth}
\raggedright\footnotesize Rows are ranked by article count, then by source count.
\end{minipage}%
}
\end{table*}

\section{Discussion}

The comparison with GDELT should be interpreted with care. GDELT is not a human-labeled gold standard, but another automated annotation system. Therefore, differences between our results and GDELT do not automatically mean that one side is wrong. The moderate agreement scores show that the two systems often overlap, but also capture different parts of the text. In particular, the LLM found many person mentions that were not present in the GDELT output, which is useful for bias analysis because missing a public figure can weaken later source and tone comparisons.

The prompt comparison also shows a practical trade-off. Prompt v2 reduced inconsistencies between sentiment labels and scores, but it also processed fewer articles and took longer to run. This suggests that adding strict validation rules directly to the prompt can make the task harder for a local model. For a continuous pipeline, Prompt v1 is therefore more practical: it produces broader coverage, runs faster, and remaining inconsistencies can be handled afterward with simple rule-based checks.

Another useful benefit of LLM-based annotation is that the model helps clean up messy news data. In multilingual and low-resource situations, raw article text often has inconsistent names, different spellings, formatting errors, and mixed language forms. A traditional process would need a lot of manual cleaning before entity and sentiment analysis could be done reliably. In contrast, the LLM can often turn these surface differences into more consistent entity and sentiment results, lowering the amount of preparation needed before analysis.

The person-tone results should also be read with caution. Large tone differences across sources may reflect real editorial framing, but they may also be affected by biases already present in the model. This is especially important for political figures, where the model may have learned associations before seeing the article being analyzed. Future versions of the pipeline should test mitigation strategies such as replacing politician names with neutral placeholders \cite{elbouanani_analyzing_2025} and adding human feedback for disputed cases.

\section{Conclusion}

In this paper, we presented an end-to-end NLP pipeline that uses local LLMs and the Kedro framework to track media bias, framing, and gatekeeping in online news. We applied the pipeline to 8,358 Albanian news articles from April 2026 and showed that a general-purpose open-weights model such as Gemma 4 can extract useful semantic signals in a low-resource setting without a large local training corpus. The evaluation showed that stricter prompt-level validation improves label-score consistency and JSON schema compliance, but reduces processing speed and annotation coverage. For this reason, the simpler prompt, combined with external rule-based data cleaning, is the more practical choice for continuous ingestion pipelines.

Given the relatively low coverage rates and the computational cost of extracting NER and sentiment information from news articles, future work will train lighter-weight machine-learning models using LLM annotations and crowdsourced data.

The media-bias and framing analysis in Section \ref{subsec:media_bias_framing_analysis} shows how source-level profiles summarize general framing tendencies, person-level tone balances show how public figures are treated across outlets, and event-level indicators show how widely stories are covered. Taken together, these views provide a practical basis for comparing news sources and for building tools that help readers inspect coverage differences more easily.

These results also show the practical role of LLM-based enrichment in a news-analysis workflow. The goal is not to replace editorial judgment or expert review, but to reduce the amount of material that must be inspected manually by turning large article collections into clearer signals. This is particularly useful in low-resource media settings, where continuous monitoring is difficult and language-specific NLP tools are still limited.

For this reason, LLM annotations are best understood here as a fast bootstrapping layer rather than as a state-of-the-art alternative to human validation. They provide a useful baseline from which similar media-bias analyses can be started quickly, especially when no manually normalized dataset is available. Since most people mentioned in news articles are public figures, a later human editorial validation step is also feasible: editors can review a relatively constrained set of extracted names, aliases, and disputed tone assignments instead of annotating the full article collection from scratch.

In addition to providing dashboard visualizations, we plan to create an MCP server that would make the media-bias and framing functions available to AI-powered systems. We believe that a continuous media-bias monitoring pipeline is most useful when integrated into an existing system, such as a news aggregator, where it can show readers polarization indicators or surface events covered from different angles by other news providers.

\bibliographystyle{IEEEtran}
\bibliography{references}

@inproceedings{ye_mediarank_2019,
	address = {Anchorage AK USA},
	title = {{MediaRank}: {Computational} {Ranking} of {Online} {News} {Sources}},
	isbn = {978-1-4503-6201-6},
	shorttitle = {{MediaRank}},
	url = {https://dl.acm.org/doi/10.1145/3292500.3330709},
	doi = {10.1145/3292500.3330709},
	language = {en},
	urldate = {2026-03-30},
	booktitle = {Proceedings of the 25th {ACM} {SIGKDD} {International} {Conference} on {Knowledge} {Discovery} \& {Data} {Mining}},
	publisher = {ACM},
	author = {Ye, Junting and Skiena, Steven},
	month = jul,
	year = {2019},
	pages = {2469--2477},
}

@article{castillo-campos_automated_2025,
	title = {Automated {Detection} of {Media} {Bias} {Using} {Artificial} {Intelligence} and {Natural} {Language} {Processing}: {A} {Systematic} {Review}},
	issn = {0894-4393, 1552-8286},
	shorttitle = {Automated {Detection} of {Media} {Bias} {Using} {Artificial} {Intelligence} and {Natural} {Language} {Processing}},
	url = {https://journals.sagepub.com/doi/10.1177/08944393251331510},
	doi = {10.1177/08944393251331510},
	language = {en},
	urldate = {2026-03-30},
	journal = {Social Science Computer Review},
	author = {Castillo-Campos, Mar and Becerra-Alonso, David and Boomgaarden, Hajo G.},
	month = apr,
	year = {2025},
	pages = {08944393251331510},
}

@article{tohidi_rethinking_2025,
	title = {Rethinking news framing with large language models},
	volume = {15},
	copyright = {2025 The Author(s)},
	issn = {2045-2322},
	url = {https://www.nature.com/articles/s41598-025-29519-9},
	doi = {10.1038/s41598-025-29519-9},
	language = {en},
	number = {1},
	urldate = {2026-03-29},
	journal = {Scientific Reports},
	publisher = {Nature Publishing Group},
	author = {Tohidi, Amir and Haider, Samar and Watts, Duncan J.},
	month = nov,
	year = {2025},
	pages = {45592},
}

@inproceedings{wang_media_2025,
	address = {Yokohama Japan},
	title = {Media {Bias} {Detector}: {Designing} and {Implementing} a {Tool} for {Real}-{Time} {Selection} and {Framing} {Bias} {Analysis} in {News} {Coverage}},
	isbn = {979-8-4007-1394-1},
	shorttitle = {Media {Bias} {Detector}},
	url = {https://dl.acm.org/doi/10.1145/3706598.3713716},
	doi = {10.1145/3706598.3713716},
	booktitle = {Proceedings of the 2025 {CHI} {Conference} on {Human} {Factors} in {Computing} {Systems}},
	language = {en},
	urldate = {2026-03-29},
	publisher = {ACM},
	author = {Wang, Jenny S and Haider, Samar and Tohidi, Amir and Gupta, Anushkaa and Zhang, Yuxuan and Callison-Burch, Chris and Rothschild, David and Watts, Duncan J},
	month = apr,
	year = {2025},
	pages = {1--27},
}

@inproceedings{r_sentinel_2025,
	address = {Sathyamangalam, India},
	title = {Sentinel: {An} {Integrated} {Framework} for {News} {Sentiment} {Analysis}, {Bias} {Detection}, and {Coverage} {Comparison} {Using} {LLMs}},
	copyright = {https://doi.org/10.15223/policy-029},
	isbn = {979-8-3503-5753-0},
	shorttitle = {Sentinel},
	url = {https://ieeexplore.ieee.org/document/11019897/},
	doi = {10.1109/STCR62650.2025.11019897},
	urldate = {2026-03-31},
	booktitle = {2025 {Fourth} {International} {Conference} on {Smart} {Technologies}, {Communication} and {Robotics} ({STCR})},
	publisher = {IEEE},
	author = {R, Prasanna Kumar and Mohan G, Bharathi and S, Akilesh Rao and Y, Jaikanth},
	month = may,
	year = {2025},
	pages = {1--6},
}

@article{ronnback_automatic_2025,
	title = {Automatic large-scale political bias detection of news outlets},
	volume = {20},
	issn = {1932-6203},
	url = {https://journals.plos.org/plosone/article?id=10.1371/journal.pone.0321418},
	doi = {10.1371/journal.pone.0321418},
	language = {en},
	number = {5},
	urldate = {2026-03-31},
	journal = {PLOS ONE},
	publisher = {Public Library of Science},
	author = {Rönnback, Ronja and Emmery, Chris and Brighton, Henry},
	month = may,
	year = {2025},
	pages = {e0321418},
}

@inproceedings{elbouanani_analyzing_2025,
	address = {Vienna, Austria},
	title = {Analyzing {Political} {Bias} in {LLMs} via {Target}-{Oriented} {Sentiment} {Classification}},
	isbn = {979-8-89176-256-5},
	url = {https://aclanthology.org/2025.findings-acl.799/},
	doi = {10.18653/v1/2025.findings-acl.799},
	urldate = {2026-03-31},
	booktitle = {Findings of the {Association} for {Computational} {Linguistics}: {ACL} 2025},
	publisher = {Association for Computational Linguistics},
	author = {Elbouanani, Akram and Dufraisse, Evan and Popescu, Adrian},
	editor = {Che, Wanxiang and Nabende, Joyce and Shutova, Ekaterina and Pilehvar, Mohammad Taher},
	month = jul,
	year = {2025},
	pages = {15476--15505},
}

@inproceedings{wessel_llm-based_2025,
	address = {Albuquerque, New Mexico},
	title = {{LLM}-based {Adversarial} {Dataset} {Augmentation} for {Automatic} {Media} {Bias} {Detection}},
	isbn = {979-8-89176-241-1},
	url = {https://aclanthology.org/2025.latechclfl-1.3/},
	doi = {10.18653/v1/2025.latechclfl-1.3},
	urldate = {2026-03-31},
	booktitle = {Proceedings of the 9th {Joint} {SIGHUM} {Workshop} on {Computational} {Linguistics} for {Cultural} {Heritage}, {Social} {Sciences}, {Humanities} and {Literature} ({LaTeCH}-{CLfL} 2025)},
	publisher = {Association for Computational Linguistics},
	author = {Wessel, Martin},
	editor = {Kazantseva, Anna and Szpakowicz, Stan and Degaetano-Ortlieb, Stefania and Bizzoni, Yuri and Pagel, Janis},
	month = may,
	year = {2025},
	pages = {19--24},
}

@inproceedings{hoxha_real_2026,
	address = {Kaunas, Lituania},
	title = {Real {Time} {Media} {Bias} and {Framing} {Detection} using {LLMs}},
	booktitle = {Book of Abstracts Scientific-practical Conference Innovations in Publishing, Printing and Multimedia Technologies 2026},
	author = {Hoxha, Klesti},
	year = {2026},
}

@article{raza_news_2022,
	title = {News recommender system: a review of recent progress, challenges, and opportunities},
	volume = {55},
	issn = {1573-7462},
	shorttitle = {News recommender system},
	url = {https://doi.org/10.1007/s10462-021-10043-x},
	doi = {10.1007/s10462-021-10043-x},
	language = {en},
	number = {1},
	urldate = {2026-05-26},
	journal = {Artificial Intelligence Review},
	author = {Raza, Shaina and Ding, Chen},
	month = jan,
	year = {2022},
	pages = {749--800},
}

@article{lazer_science_2018,
	title = {The science of fake news},
	volume = {359},
	url = {https://www.science.org/doi/abs/10.1126/science.aao2998},
	doi = {10.1126/science.aao2998},
	number = {6380},
	urldate = {2026-05-26},
	journal = {Science},
	publisher = {American Association for the Advancement of Science},
	author = {Lazer, David M. J. and Baum, Matthew A. and Benkler, Yochai and Berinsky, Adam J. and Greenhill, Kelly M. and Menczer, Filippo and Metzger, Miriam J. and Nyhan, Brendan and Pennycook, Gordon and Rothschild, David and Schudson, Michael and Sloman, Steven A. and Sunstein, Cass R. and Thorson, Emily A. and Watts, Duncan J. and Zittrain, Jonathan L.},
	month = mar,
	year = {2018},
	pages = {1094--1096},
}

@article{qin_large_2026,
	title = {Large language models meet {NLP}: a survey},
	volume = {20},
	issn = {2095-2236},
	shorttitle = {Large language models meet {NLP}},
	url = {https://doi.org/10.1007/s11704-025-50472-3},
	doi = {10.1007/s11704-025-50472-3},
	language = {en},
	number = {11},
	urldate = {2026-05-26},
	journal = {Frontiers of Computer Science},
	author = {Qin, Libo and Chen, Qiguang and Feng, Xiachong and Wu, Yang and Zhang, Yongheng and Li, Yinghui and Li, Min and Che, Wanxiang and Yu, Philip S.},
	month = mar,
	year = {2026},
	pages = {2011361},
}

@inproceedings{leetaru2013gdelt,
  author    = {Leetaru, Kalev and Schrodt, Philip A.},
  title     = {{GDELT}: Global Data on Events, Location, and Tone, 1979--2012},
  booktitle = {Proceedings of the International Studies Association Annual Convention},
  pages     = {April 2013},
  year      = {2013},
  address   = {San Diego, CA}
}

@incollection{barolli_ner_2025,
	address = {Cham},
	title = {{NER} for {Albanian} {Language}: {A} {Manually} {Annotated} {Corpus} and {Machine} {Learning} {Models}},
	volume = {247},
	isbn = {978-3-031-87768-1 978-3-031-87769-8},
	shorttitle = {{NER} for {Albanian} {Language}},
	url = {https://link.springer.com/10.1007/978-3-031-87769-8_14},
	doi = {10.1007/978-3-031-87769-8_14},
	language = {en},
	urldate = {2026-05-30},
	booktitle = {Advanced {Information} {Networking} and {Applications}},
	publisher = {Springer Nature Switzerland},
	author = {Kote, Nelda and Kalliri, Klea and Kalliri, Kejsi and Haveriku, Alba and Muraku, Besjana and Meçe, Elinda Kajo},
	editor = {Barolli, Leonard},
	year = {2025},
	note = {Series Title: Lecture Notes on Data Engineering and Communications Technologies},
	pages = {153--165},
}

@article{kadriu_human-annotated_2022,
	title = {Human-annotated dataset for social media sentiment analysis for {Albanian} language},
	volume = {43},
	issn = {23523409},
	url = {https://linkinghub.elsevier.com/retrieve/pii/S2352340922006333},
	doi = {10.1016/j.dib.2022.108436},
	language = {en},
	urldate = {2026-05-30},
	journal = {Data in Brief},
	author = {Kadriu, Fatbardh and Murtezaj, Doruntina and Gashi, Fatbardh and Ahmedi, Lule and Kurti, Arianit and Kastrati, Zenun},
	month = aug,
	year = {2022},
	pages = {108436},
}

@inproceedings{nuci_roberta_2024,
	address = {Torino, Italia},
	title = {{RoBERTa} {Low} {Resource} {Fine} {Tuning} for {Sentiment} {Analysis} in {Albanian}},
	url = {https://aclanthology.org/2024.lrec-main.1233/},
	urldate = {2026-05-30},
	booktitle = {Proceedings of the 2024 {Joint} {International} {Conference} on {Computational} {Linguistics}, {Language} {Resources} and {Evaluation} ({LREC}-{COLING} 2024)},
	publisher = {ELRA and ICCL},
	author = {Nuci, Krenare Pireva and Landes, Paul and Di Eugenio, Barbara},
	editor = {Calzolari, Nicoletta and Kan, Min-Yen and Hoste, Veronique and Lenci, Alessandro and Sakti, Sakriani and Xue, Nianwen},
	month = may,
	year = {2024},
	pages = {14146--14151},
}

\end{document}